%% file: paper.tex
\documentclass[10pt,draftcls,onecolumn]{IEEEtran}

\usepackage{times}
\usepackage{hyperref}
\usepackage{url}
\usepackage{bbold}
\usepackage{amsfonts}
\usepackage{amsmath,amssymb}
\usepackage{cite} 
\usepackage{graphicx}
\usepackage{hyperref}
\usepackage{wrapfig}
\usepackage{times}
\usepackage{amsthm}
\usepackage{mathrsfs} 
\usepackage{algorithm,algorithmic}
\usepackage{array}

\include{header}

\newcommand{\Real}{\mathbb{R}}

\newcommand\numberthis{\addtocounter{equation}{1}\tag{\theequation}}
\newcommand{\R}{\mathbf{Reg}}

\newcommand{\Rz}{\mathcal{R}}
\newcommand{\Dr}{\mathcal{D}_{\mathcal{R}}}
\renewcommand{\Reg}{\mathbf{Reg}^d}

\renewcommand{\F}{\mathcal{F}}
\renewcommand{\X}{\mathcal{X}}
\newcommand{\His}{\mathcal{H}}
\renewcommand{\x}{\hat{x}}

\begin{document}





\title{Online Optimization : Competing with Dynamic Comparators}

\author{Ali Jadbabaie$^1$, Alexander Rakhlin$^2$, Shahin Shahrampour$^1$ and Karthik Sridharan$^3$\footnote{

\noindent 
$[1]$ Ali Jadbabaie and Shahin Shahrampour are with the Department of Electrical and Systems Engineering at the University of Pennsylvania, Philadelphia, PA 19104 USA. (e-mail: shahin@seas.upenn.edu; jadbabai@seas.upenn.edu).\\
$[2]$ Alexander Rakhlin is with the Department of Statistics at the University of Pennsylvania, Philadelphia, PA 19104 USA. (e-mail: rakhlin@wharton.upenn.edu).\\
$[3]$ Karthik Sridharan is with the Department of Computer Science at Cornell University, Ithaca, NY 14850 USA. (e-mail: sridharan@cs.cornell.edu).}} %

\maketitle

\begin{abstract}
	Recent literature on online learning has focused on developing adaptive algorithms that take advantage of a {\it regularity} of the sequence of observations, yet retain worst-case performance guarantees. A complementary direction is to develop prediction methods that perform well against complex benchmarks. In this paper, we address these two directions together. We present a fully adaptive method that competes with dynamic benchmarks in which regret guarantee scales with regularity of the sequence of cost functions and comparators. Notably, the regret bound adapts to the smaller complexity measure in the problem environment. Finally, we apply our results to drifting zero-sum, two-player games where both players achieve no regret guarantees against best sequences of actions in hindsight.	
\end{abstract}

\input{intro}

\input{prelim}

\input{adaptiveomd}

\input{app}

\input{conclusion}

\input{supp}
\input{gameprf}

\end{document}

%% file: header.tex
\usepackage{graphicx}
\usepackage{amsmath,amsfonts,amssymb,amsthm}
\usepackage{graphicx}
\usepackage{color}

\newtheorem{theorem}{Theorem}
\newtheorem{lemma}[theorem]{Lemma}

\newtheorem{corollary}[theorem]{Corollary}

\newtheorem{proposition}[theorem]{Proposition}

\theoremstyle{definition}

\let\inf\undef
\DeclareMathOperator*{\inf}{\vphantom{p}inf}
\let\sup\undef
\DeclareMathOperator*{\sup}{\vphantom{p}sup}

\newcommand{\mc}[1]{\mathcal{#1}}
\newcommand{\mrm}[1]{\mathrm{#1}}

\newcommand{\norm}[1]{\left\|#1\right\|}

\newcommand{\argmin}[1]{\underset{#1}{\mrm{argmin}} \ }

\newcommand{\inner}[1]{\left\langle #1 \right\rangle}

\newcommand{\ind}[1]{{\bf 1}\left\{#1\right\}}
\newcommand{\tr}{\ensuremath{{\scriptscriptstyle\mathsf{T}}}}

\newcommand\x{\mathbf{x}}

\newcommand\cD{\mathcal{D}}

\newcommand\cR{\mathcal{R}}
\newcommand\X{\mathcal{X}}

\newcommand\F{\mathcal{F}}

\newcommand\Reg{\mbf{Reg}}



%% file: intro.tex
\section{Introduction}

The focus of this paper is an online optimization problem in which a {\it learner} plays against an {\it adversary} or {\it nature}. At each round $t \in \{1,\ldots,T\}$, the learner chooses an action $x_t$ from some convex feasible set $\X \subseteq \Real^d$. Then, nature reveals a convex function $f_t\in \F$ to the learner. As a result, the learner incurs the corresponding {\it loss} $f_t(x_t)$. A learner aims to minimize his {\it regret}, a comparison to a single best action in hindsight: 
\begin{align}\label{S Regret}
\R^s_T \triangleq \sum_{t=1}^Tf_t(x_t)-\min_{x \in \X} \sum_{t=1}^Tf_t(x).
\end{align}
Let us refer to this as {\it static} regret in the sense that the comparator is {\it time-invariant}. In the literature, there are numerous algorithms that guarantee a static regret rate of $\mathcal{O}(\sqrt{T})$  (see e.g. \cite{cesa2006prediction,freund1997decision,zinkevich2003online}). Moreover, when the loss functions are strongly convex, a rate of $\mathcal{O}(\log T)$ could be achieved \cite{hazan2007logarithmic}. Furthermore, minimax optimality of algorithms with respect to the worst-case adversary has been established (see e.g. \cite{abernethy2008optimal}). 

There are two major directions in which the above-mentioned results can be strengthened: (1) by exhibiting algorithms that compete with non-static comparator sequences (that is, making the benchmark harder), and (2) by proving regret guarantees that take advantage of {\it niceness} of nature's sequence (that is, exploiting some non-adversarial quality of nature's moves). Both of these distinct directions are important avenues of investigation. In the present paper, we attempt to address these two aspects by developing a single, adaptive algorithm with a regret bound that shows the interplay between the difficulty of the comparison sequence and niceness of the sequence of nature's moves. 

With respect to the first aspect, a more stringent benchmark is a {\it time-varying} comparator, a notion that can be termed {\it dynamic} regret \cite{zinkevich2003online,bousquet2003tracking,cesa2012new,buchbinder2012unified}:
\begin{align}\label{Regret}
\Reg_T\triangleq \sum_{t=1}^Tf_t(x_t)- \sum_{t=1}^T f_t(x^\ast_t),
\end{align}
where $x^\ast_t\triangleq \text{argmin}_{x \in \X}f_t(x)$. More generally, dynamic regret against a comparator sequence $\{u_t\}_{t=1}^T$
is 
$$\Reg_T(u_1,\ldots,u_T) \triangleq \sum_{t=1}^Tf_t(x_t)- \sum_{t=1}^T f_t(u_t).$$
It is well-known that in the worst case, obtaining a bound on dynamic regret is not possible. However, it is possible to achieve worst-case bounds in terms of
\begin{align}\label{Regularity}
{C}_T (u_1,\ldots,u_T)\triangleq\sum_{t=1}^T\big\|u_t-u_{t-1} \big\|,
\end{align}
i.e., the {\it regularity} of the comparator sequence, interpolating between the static and dynamic regret notions. Furthermore, the authors in \cite{hall2013online} introduce an algorithm which proposes a variant of $C_T$ involving a dynamical model. 

In terms of the second direction, there are several ways of incorporating potential regularity of nature's sequence. The authors in \cite{rakhlin2013online,rakhlin2013optimization} bring forward the idea of predictable sequences -- a generic way to incorporate some external knowledge about the gradients of the loss functions. Let $\{M_t\}_{t=1}^T$ be a {\it predictable} sequence computable by the learner at the beginning of round $t$. This sequence can then be used by an algorithm in order to achieve regret in terms of 
\begin{align}\label{Deviation}
{D}_T\triangleq\sum_{t=1}^T \big\|\nabla f_t(x_t)-M_t\big\|_\ast^2.
\end{align}
The framework of predictable sequences captures {\it variation} and {\it path-length} type regret bounds (see e.g. \cite{hazan2010extracting,chiang2012online}). Yet another way in which niceness of the adversarial sequence can be captured is through a notion of {\it temporal variability} studied in \cite{besbes2013non}:
\begin{align}\label{Variation}
{V}_T\triangleq\sum_{t=1}^T \sup_{x \in \X} \big|f_t(x)-f_{t-1}(x)\big|.
\end{align}

What is interesting---and intuitive---dynamic regret against the optimal sequence $\{x^*_t\}_{t=1}^T$ becomes a feasible objective when $V_T$ is {\it small}. When only noisy versions of gradients are revealed to the algorithm, Besbes et al. in \cite{besbes2013non} show that using a restarted {\it Online Gradient Descent (OGD)}\cite{zinkevich2003online} algorithm, one can get a bound of form $T^{2/3}(V_T + 1)^{1/3}$ on the expected regret. However, the regret bounds attained in \cite{besbes2013non} are only valid when an upper bound on $V_T$ is known to the learner before the game begins. For the full information online convex optimization setting, when one receives exact gradients instead of noisy gradients, a bound of order $V_T$ is trivially obtained by simply playing (at each round) the minimum of the previous round. 

The three quantities we just introduced --- $C_T, D_T, V_T$ --- measure distinct aspects of the online optimization problem, and  their interplay is an interesting object of study. Our first contribution is to develop a fully adaptive method (without prior knowledge of these quantities) whose dynamic regret is given in terms of these three complexity measures. This is done for the full information online convex optimization setting, and augments the existing regret bounds in the literature which focus on only one of the three notions --- $C_T, D_T, V_T$ --- (and not all the three together). To establish a sub-linear bound on the dynamic regret, we utilize a variant of the {\it Optimistic Mirror Descent (OMD)} algorithm \cite{rakhlin2013online}. 

When noiseless gradients are available and we can calculate
variations at each round, we not only establish a regret
bound in terms of $V_T$ and $T$ (without a priori knowledge
of a bound on $V_T$ ), but also show how the bound can in fact be improved when deviation $D_T$ is $o(T)$. We further also show how the bound can automatically adapt to $C_T$ the length of sequence of comparators. Importantly, this avoids suboptimal bounds derived only in terms of one of the quantities --- $C_T,V_T$ ---  in an environment where the other one is small. 

The second contribution of this paper is the technical analysis of the algorithm. The bound on the dynamic regret is derived by applying the {\it doubling trick} to a non-monotone quantity which results in a non-monotone step size sequence (which has not been investigated to the best of authors' knowledge).


%

We provide uncoupled strategies for two players playing a sequence of drifting zero sum games. We show how when the two players play the provided strategies, their pay offs converge to the average minimax value of the sequence of games (provided the games drift slowly). In this case, both players simultaneously enjoy no regret guarantees against best sequences of actions in hindsight that vary slowly. This is a generalization of the results by Daskalakis {\it et al.} \cite{daskalakis2014near}, and Rakhlin {\it et al.} \cite{rakhlin2013optimization}, both of which are for fixed games played repeatedly.






%% file: prelim.tex

\section{Preliminaries and Problem Formulation}

\subsection{Notation}
Throughout the paper, we assume that for any action $x\in \X\subset \Real^d$ at any time $t$, it holds that
\begin{align}\label{Bdd}
|f_t(x)| \leq G.  
\end{align}
We denote by $\|\cdot \|_\ast$ the dual norm of $\|\cdot \|$, by $[T]$ the set of natural numbers $\{1,\ldots,T\}$, and by $f_{1:t}$ the shorthand of $f_1,...,f_t$, respectively.
Whenever $C_T$ is written without arguments, it will refer to regularity $C_T(x^\ast_1,\ldots,x^\ast_T)$ of the sequence of minimizers of the loss functions. We point out that our initial statements hold for the regularity of any sequence of comparators. However, for upper bounds involving $\sqrt{C_T}$, one needs to choose a computable quantity to tune the step size, and hence our main results are stated for $C_T(x^\ast_1,\ldots,x^\ast_T)$.

The quantity $D_T$ is defined with respect to an arbitrary predictable sequence $\{M_t\}_{t=1}^T$, but this dependence is omitted for brevity. 

\subsection{Comparing with existing regret bounds in the dynamic setting}
We state and discuss relevant results from the literature on online learning in dynamic environments. For any comparator sequence $\{u_t\}_{t=1}^T$ and the specific minima sequence $\{x^\ast_t\}_{t=1}^T$ the following results are  established in the literature:
\begin{center}
  \begin{tabular}{| c | c |    }
    \hline
      Reference       & Regret Notion  \\ \cline{2-2} &  Regret Rate \\ \hline 
     \cite{zinkevich2003online},\cite{hall2013online}    & $\sum_{t=1}^Tf_t(x_t)-f_t(u_t)$ \\ \cline{2-2} 
           &  $\mathcal{O}\left(\sqrt{T}(1+{C}_T(u_1,\ldots,u_T))\right)$ \\ \hline  
     \cite{besbes2013non}   & $\sum_{t=1}^T\mathbb{E}\left[f_t(x_t)\right]- f_t(x^\ast_t)$    \\ \cline{2-2}
     & $\mathcal{O}\left(T^{2/3}(1+{V}_T)^{1/3}\right)$ \\ \hline
     \cite{rakhlin2013optimization} & $\sum_{t=1}^Tf_t(x_t)- f_t(u)$ \\ \cline{2-2}
     & $\mathcal{O}\left(\sqrt{{D}_T}\right)$ \\ \hline
     Our work & $\sum_{t=1}^Tf_t(x_t)- f_t(x^\ast_t)$ \\ \cline{2-2}
     & $\tilde{\mathcal{O}}\left(\sqrt{{D}_T+1} +  \min\left\{ \sqrt{({D}_{T}+1) {C}_{T}} ,  ({D}_{T}+1)^{1/3} T^{1/3} {V}^{1/3}_{T}\right\}\right)$ \\ \hline
    \end{tabular}
\end{center}
where $\tilde{\mathcal{O}}(\cdot)$ hides the $\log T$ factor. Lemma~\ref{Sasha-Karthik} below also yields a rate of {\small$\mathcal{O}\left(\sqrt{D_T+1}(1+{C}_T(u_1,\ldots,u_T))\right)$} for any comparator sequence $\{u_t\}_{t=1}^T$. A detailed explanation of the bounds will be done after Theorem~\ref{Main Theorem}.

We remark that the authors in \cite{besbes2013non} consider a setting in which a {\it variation budget} (an upper bound on $V_T$) is known to the learner, but he/she only has noisy gradients available. Then, the restarted {\bf OGD} guarantees the mentioned rate for convex functions; the rate is modified to $\sqrt{(V_T+1)T}$ for strongly convex functions. 

For the case of \emph{noiseless} gradients, we first aim to show that our algorithm is adaptive in the sense that the learner needs not know an upper bound on $V_T$ in advance when he/she
can calculate variations observed so far. Furthermore, we shall establish that our method recovers the known bounds for stationary settings (as well as cases where $V_T$ does not change gradually along the time horizon)




\subsection{Comparison of Regularity and Variability}

We now show that $V_T$ and $C_T$ are not comparable in general.
To this end, we consider the classical problem of prediction with expert advice. In this setting, the learner deals with the linear loss $f_t(x)=\left<f_t,x\right>$ on the $d$-dimensional probability simplex. Assume that for any $t\geq 1$, we have the vector sequence 
\begin{align*}
f_t=\left\{\begin{array}{rllc} (-\frac{1}{T},0,0,\ldots,0)&,\mbox{ if } \ \ &t \  \text{even}\\
(0,-\frac{1}{T},0,\ldots,0)&, \mbox{ if } \ \  &t \ \text{odd} \end{array}\right. .
\end{align*}
Setting $u_t$, the comparator of round $t$, to be the minimizer of $f_t$, i.e. $u_t=x^\ast_t$, we have
\begin{align*}
{C}_T&=\sum_{t=1}^T \| x^\ast_t- x^\ast_{t-1}\|_1=\Theta(T) \ \ \ \ \ \ \ \   V_T =\sum_{t=1}^T \| f_t- f_{t-1}\|_\infty = \mathcal{O}\left(1\right),
\end{align*}
according to \eqref{Regularity} and \eqref{Variation}, respectively. We see that $V_T$ is considerably smaller than $C_T$ in this scenario. 
On the other hand, consider prediction with expert advice with two experts. Let $f_t=(-1/2,0)$ on even rounds and $f_t=(0,1/2)$ on odd rounds. Expert 1 remains to be the best throughout the game, and thus $C_T=\mathcal{O}(1)$, while variation $V_T = \Theta(T)$. Therefore, one can see that taking into account only one measure might lead us to suboptimal regret bounds. We show that both measures play a key role in our regret bound. Finally, we note that if $M_t = \nabla f_{t-1}(x_{t-1})$, the notion of $D_T$ can be related to $V_T$ in certain cases, yet we keep the predictable sequence arbitrary and thus as playing a role separate from $V_T$ and $C_T$.

%% file: adaptiveomd.tex

\section{Main Results}

\subsection{Optimistic Mirror Descent and Relation to Regularity}
We now outline the {\bf OMD} algorithm previously proposed in \cite{rakhlin2013online}. Let $\Rz$ be a 1-strongly convex function with respect to a norm $\|\cdot \|$, and $\Dr(\cdot,\cdot)$ represent the Bregman divergence with respect to $\Rz$. Also, let $\His_t$ be the set containing all available information to the learner at the beginning of time $t$. Then, the learner can compute the vector $M_t : \His_{t} \to \Real^d$, which we call the predictable process. Supposing that the learner has access to the side information $M_t \in \Real^d$ from the outset of round $t$, the {\bf OMD} algorithm is characterized via the following interleaved sequence,
\begin{align}
x_{t}&=\text{argmin}_{x \in \X} \bigg\{ \eta_t \big<x , M_t\big> + \Dr(x,\x_{t-1}) \bigg\} \label{Action 1}\\
\x_{t}&=\text{argmin}_{x \in \X} \bigg\{ \eta_t \big<x , \nabla_t\big> + \Dr(x,\x_{t-1}) \bigg\} \label{Action 2}, 
\end{align}
where $\nabla_t \triangleq \nabla f_t(x_t)$, and $\eta_t$ is the {\it step size} that can be chosen adaptively to attain low regret. One could observe that for $M_t = 0$, the {\bf OMD} algorithm amounts to the well-known {\it Mirror Descent} algorithm \cite{nemirovskici1983problem,beck2003mirror}. On the other hand, the special case of $M_t=\nabla_{t-1}$ recovers the scheme proposed in \cite{chiang2012online}. 
It is shown in \cite{rakhlin2013online} that the static regret satisfies
\begin{align*}
\R^s_T \leq 4R_{\max}\left(\sqrt{{D}_T}+1\right),
\end{align*}
using the step size 
\begin{align*}
\eta_t=R_{\max}\min\left\{\left(\sqrt{{D}_{t-1}}+\sqrt{{D}_{t-2}} \right)^{-1},1\right\},
\end{align*}
where $R^2_{\max}\triangleq \sup_{x,y \in \X} \Dr(x,y)$. The following lemma extends the result to arbitrary sequence of comparators $\{u_t\}_{t=1}^T$. Throughout, we assume that $\norm{\nabla_0 - M_{0}}^2_*=1$ by convention. 
\begin{lemma}\label{Sasha-Karthik}
Let $\X$ be a convex set in a Banach space $\mathcal{B}$. Let $\Rz : \mathcal{B} \mapsto \Real$ be a 1-strongly convex function on $\X$ with respect to a norm $\|\cdot \|$, and let $\|\cdot \|_\ast$ denote the dual norm. For any $L>0$, employing the time-varying step size
\begin{align*}
\eta_t = \frac{L}{\sqrt{\sum_{s=0}^{t-1} \norm{\nabla_s - M_s}^2_\ast} + \sqrt{\sum_{s=0}^{t-2} \norm{\nabla_s - M_s}^2_\ast}}, 
\end{align*}
and running the Optimistic Mirror Descent algorithm for any comparator sequence $\{u_t\}_{t=1}^T$, yields 
\begin{align*}
\Reg_T(u_1&,\ldots,u_T) \leq 2\sqrt{1+D_T} L+2\sqrt{1+D_T}\frac{\gamma C_T(u_1,\ldots,u_T)+4R_{\max}^2}{L}   ,
\end{align*}
so long as $\Dr(x,z)-\Dr(y,z) \leq \gamma \|x- y\|,  \forall x,y,z \in \X$.
\end{lemma}

Lemma \ref{Sasha-Karthik} underscores the fact that one can get a tighter bound for regret once the learner advances a sequence of conjectures $\{M_t\}_{t=1}^T$ well-aligned with the gradients. Moreover, if the learner has prior knowledge of $C_T$ (or an upper bound on it), then the regret bound would be $\mathcal{O}\left(\sqrt{({D}_T+1) C_T}\right)$ by tuning $L$. 

Note that when the function $\Rz$ is Lipschitz on $\X$, the Lipschitz condition on the Bregman divergence is automatically satisfied. For the particular case of KL divergence this can be achieved via mixing a uniform distribution to stay away from boundaries (see e.g. section 4.2 of the paper in this regard). In this case, the constant $\gamma$ is of $\mathcal{O}(\log T)$.

\subsection{The Adaptive Optimistic Mirror Descent Algorithm}\label{AOMD Section}

The main objective of the paper is to develop the {\it Adaptive Optimistic Mirror Descent (AOMD)} algorithm. The {\bf AOMD} algorithm incorporates all notions of variation $D_T$, $C_T$ and $V_T$ to derive a comprehensive regret bound. The proposed method builds on the {\bf OMD} algorithm with adaptive step size, combined with a {\it doubling trick} applied to a threshold growing non-monotonically (see e.g. \cite{cesa2006prediction,rakhlin2013online} for application of doubling trick on monotone quantities). The scheme is adaptive in the sense that no prior knowledge of $D_T$, $C_T$ or $V_T$ is necessary.


Observe that the prior knowledge of a variation budget (an upper bound on $V_T$) does not tell us how the changes between cost functions are distributed throughout the game. For instance, the variation can increase gradually along the time horizon, while it can also take place in the form of discrete switches. The learner does not have any information about the variation pattern. Therefore, she must adopt a flexible strategy that achieves low regret in the benign case of finite switches or shocks, while it is simultaneously able to compete with the worst-case of gradual change. Before describing the algorithm, let us first use Lemma \ref{Sasha-Karthik} to bound the general dynamic regret in terms of $D_T$, $C_T$ and $V_T$. 

\begin{lemma}\label{Major Lemma}
Let $\X$ be a convex set in a Banach space $\mathcal{B}$. Let $\Rz : \mathcal{B} \mapsto \Real$ be a 1-strongly convex function on $\X$ with respect to a norm $\|\cdot \|$. Run the Optimistic Mirror Descent algorithm with the step size given in the statement of Lemma \ref{Sasha-Karthik}. Letting the comparator sequence be $\{u_t\}_{t=1}^T$, for any $L>2R_{\max}$ we have
\begin{align*}
&\Reg_T(u_1,\ldots,u_T) \leq 4 \sqrt{1+{D}_T} L  + \ind{\gamma{C}_T(u_1,\ldots,u_T) > L^2 - 4 R^2_{\max} } \frac{4\gamma R_{\max}T V_T}{L^2 - 4 R^2_{\max}},
\end{align*}
so long as $\Dr(x,z)-\Dr(y,z) \leq \gamma\|x- y\|,  \forall x,y,z \in \X$.
\end{lemma}

We now describe {\bf AOMD} algorithm shown in table \ref{ALGO}, and prove that it automatically adapts to $V_T$, $D_T$ and $C_T$. The algorithm can be cast as a repeated {\bf OMD} using different step sizes. The learner sets the parameter $L=3R_{\max}$ in Lemma \ref{Sasha-Karthik}, and runs the {\bf OMD} algorithm. Along the process, the learner collects deviation, variation and regularity observed so far, and checks the doubling condition in table \ref{ALGO} after each round. Once the condition is satisfied, the learner doubles $L$, discards the accumulated deviation, variation and regularity, and runs a new {\bf OMD} algorithm. Note importantly that the doubling condition results in a non-monotone sequence of step size during the learning process.
\begin{algorithm}[h]
\caption{Adaptive Optimistic Mirror Descent Algorithm}
\begin{algorithmic}\label{ALGO}
\STATE Parameter : $R_{\max}$, some arbitrary $x_0 \in \X$
\STATE Initialize $N = 1$, $C_{(1)} = V_{(1)} = 0$, $D_{(1)} = 1$, $x_1 = x_0$, $L_1 = 3R_{\max}$, $\Delta_1 = 0$ and $k_1 = 1$.
\FOR{$t = 1$ to $T$} 
\STATE {\tt\%  check doubling condition }
 	\IF{$ L^2_N <  \gamma\min\left\{C_{(N)} \ , \  V^{2/3}_{(N)}  \Delta_{N}^{2/3}  D_{(N)}^{-1/3} \right\} + 4 R^2_{\max} $ }
		\STATE {\tt\% increment $N$ and double $L_N$ }
		\STATE $N = N+1$
		\STATE $L_N = 3R_{\max}2^{N-1}$, $C_{(N)} = V_{(N)} = 0$, $D_{(N)} = 1$ and $\Delta_N = 0$
		\STATE $k_N = t$
	\ENDIF
	\STATE Play $x_t$ and suffer loss $f_t(x_t)$
	\STATE Calculate $M_{t+1}$ (predictable sequence) and gradient $\nabla_{t} = \nabla f_t(x_t)$
	\STATE {\tt\% update $D_{(N)}, C_{(N)}, V_{(N)}$ and $\Delta_N$} 
	\STATE $D_{(N)} = D_{(N)} + \norm{\nabla_t - M_t}^2_*$
	\STATE $C_{(N)} = C_{(N)} + \norm{x^*_t - x^*_{t-1}}$
	\STATE $V_{(N)} = V_{(N)} + \sup_{x \in \X} |f_{t}(x) - f_{t-1}(x)|$
	\STATE $\Delta_N = \Delta_N + 1$

	\STATE {\tt\% set step-size and perform optimistic mirror descent update} 
	\STATE $\eta_{t+1} = L_N \left(\sqrt{D_{(N)}} + \sqrt{D_{(N)} - \norm{\nabla_t - M_t}^2_*}\right)^{-1}$
	\STATE $\x_{t} =\argmin{x \in \X} \bigg\{ \eta_t \big<x , \nabla_t\big> + \Dr(x,\x_{t-1}) \bigg\}$\\ $x_{t+1} = \argmin{x \in \X} \bigg\{ \eta_{t+1} \big<x , M_{t+1}\big> + \Dr(x,\x_{t}) \bigg\}$
\ENDFOR
\end{algorithmic}
\label{alg:adaplocmet}
\end{algorithm}

Notice that once we have completed running the algorithm, $N$ is the number of doubling epochs, $\Delta_i$ is the number of instances in epoch $i$, $k_i$ and $k_{i+1}-1$ are the start and end points of epoch $i$, $\sum_{i=1}\Delta_i = T$ , $\sum_{i=1}^N C_{(i)} = C_T$, $\sum_{i=1}^N D_{(i)} = D_T+N$  and $\sum_{i=1}^N V_{(i)} = V_T$. Also, there is a technical reason for initialization choice of $L$ which shall become clear in the proof of Lemma \ref{Major Lemma}. Theorem \ref{Main Theorem} shows the bound enjoyed by the proposed {\bf AOMD} algorithm.

%
%

\begin{theorem}\label{Main Theorem}
Assume that $\Dr(x,z)-\Dr(y,z) \leq \gamma\|x- y\|,  \forall x,y,z \in \X$, and let  $C_T=\sum_{t=1}^T \norm{x^*_t - x^*_{t-1}}$. The {\bf AOMD} algorithm enjoys the following bound on dynamic regret : 
\begin{align*}
&\Reg_T \leq  \tilde{\mathcal{O}}\left(  \sqrt{{D_T+1}} \right) +  \tilde{\mathcal{O}}\left( \min\left\{ \sqrt{({D}_{T}+1) {C}_{T}} ,  ({D}_{T}+1)^{1/3} T^{1/3} {V}^{1/3}_{T}\right\} \right),
\end{align*}
where $\tilde{\mathcal{O}(\cdot)}$ hides a $\log T$ factor.
\end{theorem}
Based on Theorem \ref{Main Theorem} we can obtain the following table that summarizes bounds on $\Reg_T$ for various cases (disregarding the first term $\tilde{\mathcal{O}}\left(  \sqrt{{D_T+1}} \right)$ in the bound above):

\begin{center}
  \begin{tabular}{|c||c|} 
    \hline
    Regime & Rate \\ \hline
    ${C}_T  \le T^{2/3} ({D}_T+1)^{-1/3} {V}_T^{2/3}$ &  $\tilde{\mathcal{O}}\left(\sqrt{{C}_T ({D}_T+1)}\right)$ \\ \hline
    ${V}_T \le  {D}_T+1$ & $\tilde{\mathcal{O}}\big(({D}_T+1)^{2/3} T^{1/3}\big)$ \\ \hline
    ${D}_T \le {V}_T-1$ &  $\tilde{\mathcal{O}}\big({V}_T^{2/3} T^{1/3}\big)$ \\ \hline
    ${D}_T=  \mathcal{O}(T)$ & $\tilde{\mathcal{O}}\big(T^{2/3}{V}_T^{1/3}\big)$ \\ \hline
  \end{tabular}
\end{center}

The following remarks are in order : 
\begin{itemize}
\item In all cases, given the condition $V_T=o(T)$, the regret is sub-linear. When the gradients are bounded, the regime $D_T= \mathcal{O}(T)$ always holds, guaranteeing the worst-case bound  of $\mathcal{\tilde{O}}\big(T^{2/3}V_T^{1/3}\big)$.  
\item Theorem \ref{Main Theorem} allows us to recover $\mathcal{\tilde{O}}(1)$ regret for certain cases where $V_T=\mathcal{O}(1)$. Let nature divide the horizon into $B$ batches, and play a smooth convex function $f_i(x)$ on each batch $i\in [B]$, that is for some $H_i>0$ it holds that
\begin{align}\label{smoothness}
\left\|\nabla f_i(x)-\nabla f_i(y)\right\|_* \leq H_i\|x-y\|, 
\end{align}
$\forall i\in [B] \ \ \text{and} \ \ \forall x,y \in \X.$ Set $M_t=\nabla f_i(\x_{t-1})$ and note that the gradients are Lipschitz continuous. In this case, the {\bf OMD} corresponding to each batch can be recognized as the {\it Mirror Prox} method \cite{nemirovski2004prox}, which results in $\mathcal{\tilde{O}}(1)$ regret during each period. Also, since $C_T=\mathcal{O}(1)$ the bound in Theorem \ref{Main Theorem} is of $\mathcal{O}(\log T)$. 
\end{itemize}

%% file: app.tex

\section{Applications}
\subsection{Competing with Strategies}
So far, we mainly considered dynamic regret $\Reg_T$ defined in Equation \ref{Regret}. However, in many scenarios one might want to consider regret against a more specific set of strategies, defined as follows :
$$
\R^\Pi_T \triangleq \sum_{t=1}^Tf_t(x_t)- \inf_{\pi \in \Pi} \sum_{t=1}^T f_t(\pi_t(f_{1:t-1})),
$$
where each $\pi \in \Pi$ is a sequence of mappings $\pi = (\pi_1,\ldots, \pi_T)$ and $\pi_t: \F^{t-1} \to \X$. Notice that if $\Pi$ is the set of all mappings then  $\R^\Pi_T$ corresponds to dynamic regret $\Reg_T$ and if $\Pi$ corresponds to set of constant history independent mappings, that is, each $\pi \in \Pi$ is indexed by some $x \in \X$ and $\pi^x_{1}(\cdot) = \ldots = \pi^x_T(\cdot) = x$, then $\R^\Pi_T$ corresponds to the static regret $\R^s_T$. 
We now define $${C}^{\Pi}_T =  \sum_{t=1}^T \norm{\pi^*_t(f_{1:t-1}) - \pi^*_{t-1}(f_{1:t-2})},$$ 
where $\pi^*_t = \text{arginf}_{\pi \in \Pi} \sum_{s=1}^t f_s(\pi_s(f_{1:s-1}))$. Assume that there exists sequence of mappings $\tilde{C}_1,\ldots,\tilde{C}_T$ where $\tilde{C}_t$ maps any $f_1,\ldots,f_{t}$ to reals and is such that for any $t$ and any $f_1,\ldots,f_{t-1}$,
$$
\tilde{C}_{t-1}(f_{1:t-1}) \le \tilde{C}_{t}(f_{1:t}),
$$
and further, for any $T$ and any $f_1,\ldots,f_T$, 
$$
\sum_{t=1}^T \norm{\pi^*_t(f_{1:t-1}) - \pi^*_{t -1}(f_{1:t-2})} \le \tilde{C}_{T}(f_{1:T}).
$$
In this case a simple modification of {\bf AOMD} algorithm where $C_{(N)}$'s are replaced by $\tilde{C}_{\Delta_N}(f_{k_{N}:k_{N+1}-1})$ leads to the following corollary of Theorem \ref{Main Theorem}.

\begin{corollary}
Assume that $\Dr(x,z)-\Dr(y,z) \leq \gamma \|x- y\|, \forall x,y,z \in \X$. The {\bf AOMD} algorithm with the modification mentioned above achieves the following bound on regret  
\begin{align*}
&\R^\Pi_T \leq  \tilde{\mathcal{O}}\left(  \sqrt{{D_T+1}} \right) +  \tilde{\mathcal{O}}\left( \min\left\{ \sqrt{({D}_{T}+1) \tilde{C}_{T}(f_{1:T})} ,  ({D}_{T}+1)^{1/3} T^{1/3} {V}^{1/3}_{T}\right\} \right).
\end{align*}
\end{corollary}
The corollary naturally interpolates between the static and dynamic regret. In other words, letting $\tilde{C}_{T}(f_{1:T})=0$ (which holds for constant mappings), we recover the result of \cite{rakhlin2013optimization} (up to logarithmic factors), whereas $\tilde{C}_{T}(f_{1:T})=C_T$ simply recovers the regret bound in Theorem \ref{Main Theorem} corresponding to dynamic regret. The extra log factor is the cost of adaptivity of the algorithm as we assume no prior knowledge about the environment.

\subsection{Switching Zero-sum Games with Uncoupled Dynamics}
Consider two players playing $T$ zero sum games defined by matrices $A_t \in [-1,1]^{m \times n}$ for each $t\in [T]$. We would like to provide strategies for the two players such that, if both players honestly follow the prescribed strategies, the average payoffs of the players approach the average minimax value for the sequence of games at some fast rate. Furthermore, we would also like to guarantee that if one of the players (say the second) deviates from the prescribed strategy, then the first player still has small regret against sequence of actions that do not change drastically. To this end, one can use a simple modification of the {\bf AOMD} algorithm for both players that uses KL divergence as $\mc{D}_{\mc{R}}$, and mixes in a bit of uniform distribution on each round, producing an algorithm similar to the one in \cite{rakhlin2013optimization} for unchanging uncoupled dynamic games. The following theorem provides bounds for when both players follow the strategy and bound on regret for player I when  player II deviates from the strategy.

\begin{center}
\framebox[0.82\columnwidth][c]{
    \begin{minipage}{1.2\columnwidth}
    	{\tt 
		~~~~~~~~~~~~~~~~~~On round $t$, Player I performs
				\begin{align*}
			&\text{Play} ~ x_t \text{ and observe } f_t^\top A_t \\
			&\text{Update}\\
			&~~~\x_{t}(i) \propto \x'_{t-1}(i)\exp\{-\eta_{t} [f_{t}^\tr A_t]_i \} \\ 
			&~~~\x'_{t} = \left(1-\beta\right) \x_{t} + \left(\beta/n\right){\mathbf 1}_n\\
			&~~~x_{t+1} (i) \propto \x'_{t}(i)\exp\{-\eta_{t+1} [f_{t}^\tr A_t]_i \}
		\end{align*}
 		~~~~~~~~~~~~~~~~~~and simultaneously Player II performs
						\begin{align*}
			&\text{Play} ~ f_t \text{ and observe } A_t x_t\\
			&\text{Update} \\
			&~~~\hat{f}_{t}(i) \propto \hat{f}'_{t-1}(i)\exp\{-\eta'_{t} [A_t x_{t}]_i \} \\
			&~~~\hat{f}'_{t} = \left(1-\beta\right) \hat{f}_{t} + \left(\beta/m\right){\mathbf 1}_m\\
			&~~~f_{t+1} (i) \propto \hat{f}'_{t}(i)\exp\{-\eta'_{t+1} [A_t x_{t}]_i \}
		\end{align*}
}
	\vspace{-4mm}
    \end{minipage}
}
\end{center}

Note that in the description of the algorithm as well as the following proposition and its proof, any letter with the prime symbol refers to Player II, and it is used to differentiate the letter from its counterpart for player I.

\begin{proposition}\label{lem:gamcomplex}
Define $\mathscr{F}_t\triangleq \sum_{i=1}^{t}\norm{f_i^\top A_i - f_{i-1}^\top A_{i-1}}^2_\infty$, and let 
\begin{align*}
\eta_t &= \min\left\{\log(T^2n) \frac{L }{\sqrt{\mathscr{F}_{t-1} } + \sqrt{\mathscr{F}_{t-2} } }, \frac{1}{32 L}\right\}.
\end{align*}
Also define $\mathscr{A}_t\triangleq \sum_{i=1}^{t}\norm{A_i x_i - A_{i-1} x_{i-1}}^2_\infty$, and let
\begin{align*}
\eta'_t &= \min\left\{\log(T^2m)\frac{L }{\sqrt{\mathscr{A}_{t-1} } + \sqrt{\mathscr{A}_{t-2} } }, \frac{1}{32 L}\right\}.
\end{align*} 
Let $\beta=1/T^2$, $M_t=f_{t-1}^{\top} A_{t-1}$, and $M'_t=A_{t-1}x_{t-1}$. When Player I uses the prescribed strategy, irrespective of the actions of player II, the regret of Player I w.r.t. any sequence of actions $u_1,\ldots,u_T$ is bounded as : 
\begin{align*}
\sum_{t=1}^T \left(f_t^\top A_t x_t -  f_t^\top A_t u_t\right) &\leq 2\log(T^2 n)\left( {C}_T (u_1,\ldots,u_T) + 2\right)\left(32L+\frac{2\sqrt{\mathscr{F}_T}}{\log(T^2n)L}\right)+ \log(T^2n)\frac{L}{2}\sqrt{\mathscr{F}_T}.
\end{align*}
Further if both players follow the prescribed strategies then, as long as 
\begin{align}
2L^2 >  \max\left\{ {C}_T , {C}'_T \right\}+3,\label{LCON}
\end{align}
we get, { \small
\begin{align*}
\sum_{t=1}^T \sup_{f_t \in \Delta_m} f_t^\top A_t x_t  \le \sum_{t=1}^T \inf_{x_t \in \Delta_n}\sup_{f_t \in \Delta_m} f_t^\top A_t x_t &+\frac{256L}{T}+\frac{1}{2L}+4\sum_{t=1}^T\norm{A_{t-1} - A_t}_\infty\\
& + 32L\left( \log(T^2 n) {C}_T + \log(T^2 m) C'_T  + 2 \log(T^4 n m)   \right) \\
&+\left(C_T+C'_T+4\right)\frac{20 + 4 \sqrt{\sum_{t=1}^T\norm{A_{t-1} - A_t}_\infty^2}}{L}.
\end{align*}}
\end{proposition}
A simple consequence of the above proposition is that if for instance the game matrix $A_t$ changes at most $K$ times over the $T$ rounds, and we knew this fact a priori, then by letting $L=\frac{1}{\sqrt{\log (T^2n)}}$, we get that regret for Player I w.r.t. any sequence of actions that switches at most $K$ times even when Player II deviates from the prescribed strategy is $\mathcal{O}\left((K+2) \sqrt{ \log(T^2n) T}\right)$. At the same time if both players follow the strategy, then average payoffs of the players converge to the average minimax equilibrium at the rate of $\mathcal{O}\left( L\left(K  + 2\right) \log(T^4 n m)\right)$ under the condition on $L$ given in \eqref{LCON}. This shows that if the game matrix only changes/switches a constant number of times, then players get $\sqrt{\log(T)T}$ regret bound against arbitrary sequences and comparator actions that switch at most $K$ times while simultaneously get a convergence rate of $\mathcal{O}\left(\log(T)\right)$ to average equilibrium when both players are honest. Also, when we let $K=0$ and set $L$ to some constant, the proposition recovers the rate in static setting \cite{rakhlin2013optimization} where the matrix sequence is time-invariant.

%% file: conclusion.tex
\section{Conclusion}
In this paper, we proposed an online learning algorithm for dynamic environments. We considered time-varying comparators to measure the dynamic regret of the algorithm. Our proposed method is fully adaptive in the sense that the learner needs no prior knowledge of the environment. We derive a comprehensive upper bound on the dynamic regret capturing the interplay of regularity in the function sequence versus the comparator sequence. Interestingly, the regret bound adapts to the smaller quantity among the two, and selects the best of both worlds. As an instance of dynamic regret, we considered drifting zero-sum, two-player games, and characterized the convergence rate to the average minimax equilibrium in terms of variability in the sequence of payoff matrices.

\section*{Acknowledgements}
We gratefully acknowledge the support of ONR BRC Program on Decentralized,
Online Optimization, NSF under grants CAREER DMS-0954737 and CCF-1116928, as well as Dean's
Research Fund.

%% file: supp.tex

\section*{Appendix : Proofs}

\textbf{\emph{Proof of Lemma \ref{Sasha-Karthik}}}. For any $u_t\in\X$, it holds that
	\begin{align}\label{eq:head}
		\inner{x_t-u_t, \nabla_t} = \inner{x_t-\hat{x}_{t},\nabla_t-M_{t}} + \inner{x_t-\hat{x}_{t}, M_t} + \inner{\hat{x}_{t}-u_t, \nabla_t}.
	\end{align}
	First, observe that for any primal-dual norm pair we have
	\begin{align*}
		\inner{x_t-\hat{x}_{t},\nabla_t-M_{t}} \le \norm{x_t-\hat{x}_{t}} \norm{\nabla_t - M_t}_*  \ .
        \end{align*}
	Any update of the form $a^* = \arg\min_{a\in \X} \inner{a,x} + \cD_\cR(a,c)$ satisfies for any $d\in \X$,
	\begin{align*}
		\inner{a^*-d,x} \leq  \cD_\cR(d,c)- \cD_\cR(d,a^*)-\cD_\cR(a^*,c) \ .
	\end{align*}
	This entails
	\begin{align*}
		\inner{x_t-\hat{x}_{t}, M_t} \leq \frac{1}{\eta_t} \bigg\{\cD_\cR(\hat{x}_{t},\hat{x}_{t-1}) - \cD_\cR(\hat{x}_{t},x_t) - \cD_\cR(x_t,\hat{x}_{t-1})  \bigg\} 
	\end{align*}
	and
	\begin{align*}
		\inner{\hat{x}_{t}-u_t, \nabla_t} \leq \frac{1}{\eta_t}\bigg\{\cD_\cR(u_t,\hat{x}_{t-1})  - \cD_\cR(u_t,\hat{x}_{t}) - \cD_\cR(\hat{x}_{t},\hat{x}_{t-1}) \bigg\}  \ .
	\end{align*}
	Combining the preceding relations and returning to \eqref{eq:head}, we obtain
	\begin{align*}
	\inner{x_t-u_t, \nabla_t}	&\leq  \frac{1}{\eta_t}\bigg\{\cD_\cR(u_t,\hat{x}_{t-1})  - \cD_\cR(u_t,\hat{x}_{t})  - \cD_\cR(\hat{x}_{t},x_t) - \cD_\cR(x_t,\hat{x}_{t-1}) \bigg\} \\
	 &~~~~~~~~~~~~~~~~~~~~~~~~~~+\norm{\nabla_t - M_t}_* \norm{x_t - \hat{x}_{t}} \\   
	 	&\leq \frac{1}{\eta_t}\bigg\{\cD_\cR(u_t,\hat{x}_{t-1})  - \cD_\cR(u_t,\hat{x}_{t})  - \frac{1}{2}\norm{\hat{x}_{t} - x_t}^2 - \frac{1}{2}\norm{\hat{x}_{t-1} - x_t}^2 \bigg\}\\
		&~~~~~~~~~~~~~~~~~~~~~~~~~~+\norm{\nabla_t - M_t}_* \norm{x_t - \hat{x}_{t}} , \label{eq:aistats} \numberthis
		\end{align*}
where in the last step we appealed to strong convexity: $\cD_\cR(x,y) \ge \frac{1}{2} \norm{x - y}^2$ for any $x,y \in \X$. Using the simple inequality $a b \le \frac{\rho a^2}{2} + \frac{b^2}{2\rho}$ for any $\rho>0$ to split the product term, we get
\begin{align*}
\inner{x_t-u_t, \nabla_t}	&\leq  \frac{1}{\eta_t}\bigg\{\cD_\cR(u_t,\hat{x}_{t-1})  - \cD_\cR(u_t,\hat{x}_{t})  - \frac{1}{2}\norm{\hat{x}_{t} - x_t}^2 - \frac{1}{2}\norm{\hat{x}_{t-1} - x_t}^2 \bigg\} \\
	&~~~~~~~~~~~~~~~~~~~~~~~~~~+\frac{\eta_{t+1}}{2} \norm{\nabla_t - M_t}^2_* + \frac{1}{2 \eta_{t+1}} \norm{x_t - \hat{x}_{t}}^2 , 
	\end{align*}
Applying the bound  
\begin{align*}
\frac{1}{2 \eta_{t+1}} \norm{x_t - \hat{x}_{t}}^2-\frac{1}{2 \eta_{t}} \norm{x_t - \hat{x}_{t}}^2 \leq R^2_{\max}\left(\frac{1}{\eta_{t+1}}-\frac{1}{\eta_t}\right),
\end{align*}
and summing over $t\in[T]$ yields ,
\begin{align*}
	\sum_{t=1}^T \inner{x_t-u_t, \nabla_t} 
	& \leq  \sum_{t=1}^T \frac{\eta_{t+1}}{2} \norm{\nabla_t - M_t}^2_* +  \sum_{t=1}^T \frac{1}{\eta_t}\bigg\{\cD_\cR(u_t,\hat{x}_{t-1})  - \cD_\cR(u_t,\hat{x}_{t}) \bigg\} + \frac{R_{\max}^2}{\eta_{T+1}} \\
	& \leq  \sum_{t=1}^T \frac{\eta_{t+1}}{2} \norm{\nabla_t - M_t}^2_*  +R_{\max}^2\left(\frac{1}{\eta_1} + \frac{1}{\eta_{T+1}}\right)\\
	&~~~~~~~~~~~~~~~~~~~~~~~~~~ +  \sum_{t=2}^T \bigg\{\frac{\cD_\cR(u_t,\hat{x}_{t-1})}{\eta_t} -  \frac{\cD_\cR(u_{t-1},\hat{x}_{t-1})}{\eta_{t-1}}\bigg\}\\
	&  \leq \sum_{t=2}^T \bigg\{\frac{\cD_\cR(u_t,\hat{x}_{t-1})}{\eta_t} - \frac{\cD_\cR(u_{t-1},\hat{x}_{t-1})}{\eta_t} + \frac{\cD_\cR(u_{t-1},\hat{x}_{t-1})}{\eta_t} -  \frac{\cD_\cR(u_{t-1},\hat{x}_{t-1})}{\eta_{t-1}}\bigg\}\\
	&~~~~~~~~~~~~~~~~~~~~~~~~~~+ \sum_{t=1}^T \frac{\eta_{t+1}}{2} \norm{\nabla_t - M_t}^2_*  +  \frac{2R_{\max}^2}{\eta_{T+1}}\\
	& \le  \sum_{t=1}^T \frac{\eta_{t+1}}{2} \norm{\nabla_t - M_t}^2_* +  \gamma\sum_{t=2}^T \frac{ \norm{u_t - u_{t-1}}}{\eta_t}\\
	&~~~~~~~~~~~~~~~~~~~~~~~~~~+ \sum_{t=2}^T \cD_\cR(u_{t-1},\hat{x}_{t-1}) \left(\frac{1}{\eta_t} -  \frac{1}{\eta_{t-1}}\right) +\frac{2R_{\max}^2}{\eta_{T+1}}\\
	& \le  \sum_{t=1}^T \frac{\eta_{t+1}}{2} \norm{\nabla_t - M_t}^2_* +  \gamma \sum_{t=2}^T \frac{\norm{u_t - u_{t-1}}}{\eta_t} +\frac{4R_{\max}^2}{\eta_{T+1}}\ ,
\end{align*} 
where we used the Lipschitz continuity of $\Dr$ in the penultimate step. Now let us set {\small
\begin{align*}
\eta_{t} &= \frac{L}{\sqrt{\sum_{s=0}^{t-1} \norm{\nabla_s - M_{s}}^2_*} + \sqrt{\sum_{s=0}^{t-2} \norm{\nabla_s - M_{s}}^2_\ast} }=\frac{L \left( \sqrt{\sum_{s=0}^{t-1} \norm{\nabla_s - M_{s}}^2_*} - \sqrt{\sum_{s=0}^{t-2} \norm{\nabla_s - M_{s}}^2_*}\right)}{\norm{\nabla_{t-1} - M_{t-1}}^2_*},
\end{align*}}
and $\norm{\nabla_0 - M_{0}}^2_*=1$ to have
\begin{align*}
	\sum_{t=1}^T \inner{x_t-u_t, \nabla_t} 
	& \le  \frac{L}{2} \sum_{t=1}^T \left\{ \sqrt{\sum_{s=0}^{t} \norm{\nabla_s - M_{s}}^2_*} - \sqrt{\sum_{s=0}^{t-1} \norm{\nabla_s - M_{s}}^2_*}\right\} \\
	& +  \frac{ 2\gamma\sqrt{1+\sum_{t=1}^{T} \norm{\nabla_t - M_{t}}^2_*}}{L} \sum_{t=2}^T \norm{u_t - u_{t-1}} + \frac{8 R_{\max}^2 \sqrt{1+\sum_{t=1}^{T} \norm{\nabla_t - M_{t}}^2_*}}{L}  \\
	& \le  2\sqrt{1+\sum_{t=1}^{T} \norm{\nabla_t - M_{t}}^2_*}  \left( L +  \frac{ \gamma\sum_{t=1}^T \norm{u_t - u_{t-1}} + 4 R_{\max}^2  }{L}  \right)\ .
\end{align*}
Appealing to convexity of $\{f_t\}_{t=1}^T$, and replacing $C_T$ \eqref{Regularity} and $D_T$ \eqref{Deviation} in above, completes the proof .$\hfill \blacksquare $\\

\textbf{\emph{Proof of Lemma \ref{Major Lemma}}}. We define 
\begin{align}\label{U_T}
U_T\triangleq \bigg\{u_1,...,u_T \in \X :   \gamma \sum_{t=1}^T \norm{u_t - u_{t-1}}\le L^2 - 4 R^2_{\max} \bigg\},
\end{align}
and
\begin{align*}
(u^*_1,...,u^*_T)\triangleq \text{argmin}_{u_1,\ldots,u_T \in U_T } \sum_{t=1}^T f_t(u_t).
\end{align*}
Our choice of $L>2R_{\max}$ guarantees that any sequence of fixed comparators $u_t=u$ for $t\in[T]$ belongs to $U_T$, and hence, $(u^*_1,...,u^*_T)$ exists. Noting that $(u^*_1,...,u^*_T)$ is an element of $U_T$, we have $\gamma \sum_{t=1}^T \norm{u^*_t - u^*_{t-1}}+4R^2_{\max}\leq L^2$. We now apply Lemma \ref{Sasha-Karthik} to $\{u^*_t\}_{t=1}^T$ to bound the dynamic regret for arbitrary comparator sequence $\{u_t\}_{t=1}^T$ as follows,
\begin{align*}
\Reg_T(u_1,...,u_T) &= \sum_{t=1}^T \bigg\{ f_t(x_t)-f_t(u^*_t) \bigg\} + \sum_{t=1}^T \bigg\{ f_t(u^*_t)-f_t(u_t)\bigg\}\\ 
&\le 4 \sqrt{1+D_T} L + \sum_{t=1}^T \bigg\{ f_t(u^*_t)-f_t(u_t)\bigg\}\\  
&\le 4 \sqrt{1+D_T} L + \ind{\gamma\sum_{t=1}^T \norm{u_t - u_{t-1}} > L^2 - 4 R^2_{\max} } \left( \sum_{t=1}^T \bigg\{ f_t(u^*_t)-f_t(u_t)\bigg\} \right), \numberthis \label{Product5}
\end{align*}
where the last step follows from the fact that 
\begin{align*}
\sum_{t=1}^T f_t(u^*_t)  -  \sum_{t=1}^T f_t(u_t)\leq 0  \ \ \ \ \text{if} \ \ \ \ (u_1,...,u_T) \in U_T.
\end{align*} 
Given the definition of $R^2_{\max}$, by strong convexity of $\Dr(x,y)$, we get that $\|x-y\| \leq \sqrt{2}R_{\max}$, for any $x,y \in \X$. This entails that once we divide the horizon into $B$ number of batches and use a single, fixed point as a comparator along each batch, we have
\begin{align}\label{Bcondition}
\sum_{t=1}^T \|u_t-u_{t-1}\|\leq B\sqrt{2}R_{\max},
\end{align}
since there are at most $B$ number of changes in the comparator sequence along the horizon.
Now let $B=\frac{L^2 - 4 R^2_{\max}}{\gamma\sqrt{2} R_{\max}}$, and for ease of notation, assume that $T$ is divisible by $B$.
Noting that $f_t(x^*_t)\leq f_t(u_t)$, we use an argument similar to that of \cite{besbes2013non} to get for any fixed $t_i \in [(i-1)(T/B) +1 , i(T/B)]$,
\begin{align*}
\sum_{t=1}^T \bigg\{f_t(u^*_t)-f_t(u_t)\bigg\} &\leq \sum_{t=1}^T \bigg\{f_t(u^*_t)-f_t(x^*_t)\bigg\} \numberthis\label{Product98}\\  
&=\sum_{i=1}^B\sum_{t = (i-1)(T/B) + 1}^{i(T/B)}\bigg\{f_t(u^*_t)-f_t(x^*_t) \bigg\}\\
&\leq \sum_{i=1}^B\sum_{t = (i-1)(T/B) + 1}^{i(T/B)}\bigg\{f_t(x^*_{t_i})-f_t(x^*_t)\bigg\} \numberthis \label{ECO}\\
&\leq \left(\frac{T}{B}\right) \sum_{i=1}^B \max_{t\in [(i-1)(T/B) + 1,  i (T/B)]}\bigg\{f_t(x^\ast_{t_i})-f_t(x^\ast_t)\bigg\}. \label{Product10} \numberthis
\end{align*}
Note that $x^*_{t_i}$ is fixed for each batch $i$. Substituting our choice of $B=\frac{L^2 - 4 R^2_{\max}}{\gamma\sqrt{2} R_{\max}}$ in \eqref{Bcondition} implies that the comparator sequence $u_t=x^*_{t_i}\ind{\frac{(i-1)T}{B} + 1 \leq t \leq  \frac{i T}{B}}$ belongs to $U_T$, and \eqref{ECO} follows by optimality of $(u^*_1,...,u^*_T)$. We now claim that for any $t\in [(i-1)(T/B) + 1,  i (T/B)]$, we have, 
\begin{align}
f_t(x^\ast_{t_i})-f_t(x^\ast_t)\leq 2 \sum_{s = (i-1)(T/B) + 1}^{i(T/B)} \sup_{x \in \X} |f_s(x) - f_{s-1}(x)|. \label{Product99}
\end{align}
Assuming otherwise, there must exist a $\hat{t}_i\in  [(i-1)(T/B) + 1, i (T/B)]$ such that
\begin{align*}
f_{\hat{t}_i}(x^\ast_{t_i})-f_{\hat{t}_i}(x^\ast_{\hat{t}_i})> 2 \sum_{t = (i-1)(T/B) + 1}^{i(T/B)} \sup_{x \in \X} |f_t(x) - f_{t-1}(x)|, 
\end{align*}
which results in
\begin{align*}
f_t(x^\ast_{\hat{t}_i}) & \leq f_{\hat{t}_i}(x^\ast_{\hat{t}_i})+  \sum_{t = (i-1)(T/B) + 1}^{i(T/B)} \sup_{x \in \X} |f_t(x) - f_{t-1}(x)|\\
&  < f_{\hat{t}_i}(x^\ast_{t_i})- \sum_{t = (i-1)(T/B) + 1}^{i(T/B)} \sup_{x \in \X} |f_t(x) - f_{t-1}(x)| \leq f_t(x^\ast_{t_i}),
\end{align*}
The preceding relation for $t=t_i$ violates the optimality of $x^\ast_{t_i}$, which is a contradiction. Therefore, Equation \eqref{Product99} holds for any $t\in [(i-1)(T/B) + 1,  i (T/B)]$ Combining \eqref{Product98}, \eqref{Product10} and \eqref{Product99} we have
\begin{align}\label{Product6}
\sum_{t=1}^T \bigg\{f_t(u^*_t)-f_t(u_t)\bigg\}   &\leq \frac{2 T}{B} \sum_{i=1}^B \sum_{t = (i-1)(T/B) + 1}^{i(T/B)} \sup_{x \in \X} |f_t(x) - f_{t-1}(x)| \notag\\
& =  \frac{2 T V_T}{B} =\frac{2\gamma\sqrt{2}R_{\max}TV_T}{L^2 - 4 R^2_{\max}}.
\end{align}
Using the above in Equation \eqref{Product5} we conclude the following upper bound
\begin{align*}
\Reg_T(u_1,...,u_T) \leq 4 \sqrt{1+D_T} L + \ind{\gamma\sum_{t=1}^T \norm{u_t - u_{t-1}} > L^2 - 4 R^2_{\max} } \frac{4\gamma R_{\max}T V_T}{L^2 - 4 R^2_{\max}},
\end{align*}
thereby completing the proof. $\hfill \blacksquare $\\

\textbf{\emph{Proof of Theorem \ref{Main Theorem}}}. For the sake of clarity in presentation, we stick to the following notation for the proof
\begin{align*}
\underline{D}_{(i)} &\triangleq D_{(i)}- \| \nabla_{k_{i+1} - 1} - M_{k_{i+1} - 1}\|^2_* \\
\underline{C}_{(i)} &\triangleq C_{(i)} - \|x^*_{k_{i+1} - 1}-x^*_{k_{i+1} - 2}\|\\
\underline{V}_{(i)} &\triangleq V_{(i)}- \sup_{x\in \X}\left|f_{k_{i+1} - 1}(x)-f_{k_{i+1} - 2}(x)\right|\\
\underline{\Delta}_{(i)} &\triangleq \Delta_i-1,
\end{align*}
for any doubling epoch $i=1,...,N$, where we recall that $k_{i+1}-1$ is the last instance of epoch $i$. Therefore, any symbol with lower bar refers to its corresponding quantity removing only the value of the last instance of that interval.

Let the {\bf AOMD} algorithm run with the step size given by Lemma \ref{Sasha-Karthik} in the following form 
\begin{align*}
\eta_t = \frac{L_i}{\sqrt{\sum_{s=0}^{t-1} \norm{\nabla_s - M_s}^2_\ast} + \sqrt{\sum_{s=0}^{t-2} \norm{\nabla_s - M_s}^2_\ast}}, 
\end{align*}
and let $L_i$ be tuned with a doubling condition explained in the algorithm.
Once the condition stated in the algorithm fails, the following pair of identities must hold
\begin{align}
 \gamma \min\{\underline{C}_{(i)} \ , \ \underline{\Delta}_i^{2/3} \underline{V}^{2/3}_{(i)} \underline{D}^{-1/3}_{(i)}   \} + 4 R^2_{\max}  \le L_i^2   \ \ \ \ \ \  \gamma \min\{C_{(i)} \ , \ \Delta_i^{2/3} V^{2/3}_{(i)} D^{-1/3}_{(i)}   \} + 4 R^2_{\max}  > L_i^2. \label{Pair}
\end{align}
Observe that the algorithm doubles $L_i$ only after the condition fails, so at violation points we suffer at most $2G$ by boundedness \eqref{Bdd}. Then, under purview of Lemma \ref{Major Lemma}, it holds that
\begin{align*}
\Reg_T &\le \sum_{i=1}^N\left\{ 4 \sqrt{\underline{D}_{(i)}} L_i + \ind{\gamma\underline{C}_{(i)} > L_i^2 - 4 R^2_{\max} } \frac{4\gamma R_{\max}\underline{\Delta}_i \underline{V}_{(i)}}{L_i^2 - 4 R^2_{\max}}
 \right\} + 2NG\\ \nonumber
 &\le \sum_{i=1}^N\left\{ 4 \sqrt{D_{(i)}} L_i + \ind{\underline{C}_{(i)} > \underline{\Delta}_i^{2/3} \underline{V}^{2/3}_{(i)} \underline{D}^{-1/3}_{(i)} } \frac{4\gamma R_{\max}\underline{\Delta}_i \underline{V}_{(i)}}{L_i^2 - 4 R^2_{\max}}
 \right\} + 2NG, \label{Product7} \numberthis
\end{align*}
where the last step follows directly from \eqref{Pair} and the fact that $\underline{D}_{(i)} \leq D_{(i)}$. Bounding $\sqrt{D_{(i)}}L_i$ in above, using the second inequality in \eqref{Pair}, we get 
\begin{align*}
\sqrt{D_{(i)}}L_i  &\leq \sqrt{\gamma \min\left\{D_{(i)}C_{(i)} \ , \ \Delta_i^{2/3} V^{2/3}_{(i)} D^{2/3}_{(i)}   \right\} + 4 R^2_{\max}D_{(i)}} \\
&~~~~~~~~~~~~~~~~~~~~~~~~~~\leq 2R_{\max}\sqrt{D_{(i)}} +\sqrt{\gamma}\min\left\{\sqrt{D_{(i)}C_{(i)}} \ , \ \Delta_i^{1/3} V^{1/3}_{(i)} D^{1/3}_{(i)}   \right\},
\end{align*}
by the simple inequality $\sqrt{a+b} \leq \sqrt{a}+ \sqrt{b}$. Plugging the bound above into \eqref{Product7} and noting that 
\begin{align*}
\sum_{i=1}^N \sqrt{D_{(i)}}=N\sum_{i=1}^N \frac{1}{N}\sqrt{D_{(i)}} \leq  N\sqrt{\frac{1}{N}\sum_{i=1}^ND_{(i)}}=\sqrt{ND_T+N},
\end{align*}
by Jensen's inequality, we obtain
\begin{align*}
\Reg_T & \le  2NG + 8R_{\max}  \sqrt{ND_T+N} + 4\sqrt{\gamma}\sum_{i=1}^N \min\left\{ \sqrt{D_{(i)} C_{(i)}} \ ,\  D_{(i)}^{1/3} \Delta_i^{1/3} V^{1/3}_{(i)}\right\} \\
 & ~~~~~~~~~~~~~~~~~~~~~~~~~~~~~~~+ \sum_{i=1}^N \ind{\underline{C}_{(i)} > \underline{\Delta}_i^{2/3} \underline{V}^{2/3}_{(i)} \underline{D}^{-1/3}_{(i)} } \frac{4R_{\max}\underline{\Delta}_i \underline{V}_{(i)}}{\min\left\{\underline{C}_{(i)} , \underline{\Delta}_i^{2/3} \underline{V}^{2/3}_{(i)} \underline{D}^{-1/3}_{(i)}\right\}},
 \end{align*}
 where we used the first inequality in \eqref{Pair} to bound the last term. Given the condition in the indicator function $\ind{\cdot}$, we can simplify above to derive,
 \begin{align*}
\Reg_T & \leq  2NG + 8R_{\max}  \sqrt{ND_T+N}+ 4\sqrt{\gamma}\sum_{i=1}^N \min\left\{ \sqrt{D_{(i)} C_{(i)}} \ ,\  D_{(i)}^{1/3} \Delta_i^{1/3} V^{1/3}_{(i)}\right\} \\
 & ~~~~~~~~~~~~~~~~~~~~~~~~~~~~~~~+ 4R_{\max}\sum_{i=1}^N \ind{\underline{C}_{(i)} > \underline{\Delta}_i^{2/3} \underline{V}^{2/3}_{(i)} \underline{D}^{-1/3}_{(i)} }\underline{D}^{1/3}_{(i)} \underline{V}^{1/3}_{(i)} \underline{\Delta}_i^{1/3}\\
 & =  2NG + 8R_{\max}  \sqrt{ND_T+N} + 4\sqrt{\gamma}\sum_{i=1}^N \min\left\{ \sqrt{D_{(i)} C_{(i)}} \ ,\  D_{(i)}^{1/3} \Delta_i^{1/3} V^{1/3}_{(i)}\right\} \\
 & ~~~~~~~~~~~~~~~~~~~~~~~~~~~~~~~+ 4R_{\max} \sum_{i=1}^N \ind{\sqrt{\underline{D}_{(i)}\underline{C}_{(i)}} > \underline{\Delta}_i^{1/3} \underline{V}^{1/3}_{(i)} \underline{D}^{1/3}_{(i)} }\underline{D}^{1/3}_{(i)} \underline{V}^{1/3}_{(i)} \underline{\Delta}_i^{1/3}\\
 & \le  2NG + 8R_{\max}  \sqrt{ND_T+N} + 4\sqrt{\gamma}\sum_{i=1}^N \min\left\{ \sqrt{D_{(i)} C_{(i)}} \ ,\  D_{(i)}^{1/3} \Delta_i^{1/3} V^{1/3}_{(i)}\right\} \\
 & ~~~~~~~~~~~~~~~~~~~~~~~~~~~~~~~+ 4R_{\max}\sum_{i=1}^N \min\left\{\sqrt{ \underline{D}_{(i)}\underline{C}_{(i)}}, \underline{D}^{1/3}_{(i)} \underline{V}^{1/3}_{(i)} \underline{\Delta}_i^{1/3}\right\}. \numberthis \label{Product8}
 \end{align*}
 Given the fact that 
 \begin{align*}
 \underline{C}_{(i)} \leq C_{(i)} \ \ \ \ \ \ \ \ \ \ \ \ \  \underline{D}_{(i)} \leq D_{(i)}  \ \ \ \ \ \ \ \ \ \ \ \ \  \underline{V}_{(i)} \leq V_{(i)}  \ \ \ \ \ \ \ \ \ \ \ \ \  \underline{\Delta}_i \leq \Delta_i ,
 \end{align*}
we return to \eqref{Product8} to derive
 \begin{align*}
\Reg_T & \le 2NG + 8R_{\max}  \sqrt{N D_T+N} + (4\sqrt{\gamma}+4R_{\max}) \sum_{i=1}^N \min\left\{ \sqrt{D_{(i)} C_{(i)}} \ ,\  D_{(i)}^{1/3} \Delta_i^{1/3} V^{1/3}_{(i)}\right\} \\
  & \le 2NG + 8R_{\max}  \sqrt{N D_T+N} + (4\sqrt{\gamma}+4R_{\max})  \min\left\{ \sum_{i=1}^N \sqrt{D_{(i)} C_{(i)}} , \sum_{i=1}^N D_{(i)}^{1/3} \Delta_i^{1/3} V^{1/3}_{(i)}\right\} \\
& \le 2N\left( G + 4R_{\max}  \sqrt{D_T+1} + (2\sqrt{\gamma}+2R_{\max})  \min\left\{ \sqrt{(D_{T}+1) C_{T}} ,  (D_{T}+1)^{1/3} T^{1/3} V^{1/3}_{T}\right\} \right). \label{Product9} \numberthis
\end{align*}
where we bounded the sums using the following fact about the summands
 \begin{align*}
 C_{(i)} \leq C_T \ \ \ \ \ \ \ \ \ \ \ \ \   D_{(i)} \leq D_T+1 \ \ \ \ \ \ \ \ \ \ \ \ \   V_{(i)} \leq V_T \ \ \ \ \ \ \ \ \ \ \ \ \  \Delta_i \leq T.
 \end{align*}
To bound the number of batches $N$, we recall that $L_i=3R_{\max}2^{i-1}$, and use the second inequality in \eqref{Pair} to bound $L_{N-1}$ as follows
\begin{align*}
N = 2 + \log_2(2^{N-2}) &= 2 + \log_2(L_{N-1})-\log_2(3R_{\max}) \\
& \le 2 + \frac{1}{2} \log_2\left(\gamma \min\left\{C_{(N-1)}, \Delta_{N-1}^{2/3} V^{2/3}_{(N-1)} D^{-1/3}_{(N-1)}   \right\} + 4 R^2_{\max} \right)-\log_2(3R_{\max})\\
& \le 2 + \frac{1}{2} \log_2\left(\gamma C_{(N-1)} + 4 R^2_{\max} \right)-\log_2(3R_{\max})\\
& \le 2 + \frac{1}{2} \log_2\left(2\gamma R_{\max}T + 4 R^2_{\max} \right)-\log_2(3R_{\max}). 
\end{align*}
In view of the preceding relation and \eqref{Product9}, we have 
\begin{align*}
\Reg_T \le \kappa\bigg( G + 4R_{\max}  \sqrt{D_T+1} + (2\sqrt{\gamma}+2R_{\max})  \min\left\{ \sqrt{(D_{T}+1) C_{T}} ,  (D_{T}+1)^{1/3} T^{1/3} V^{1/3}_{T}\right\} \bigg),
\end{align*}
where $\kappa\triangleq 4 +  \log_2\left(2\gamma R_{\max}T + 4 R^2_{\max} \right)-2\log_2(3R_{\max})$, thereby completing the proof. $\hfill \blacksquare $\\

\noindent

%% file: gameprf.tex

\textbf{\emph{Proof of Proposition \ref{lem:gamcomplex}}}.
Assume that the player I uses the prescribed strategy. This corresponds to using the optimistic mirror descent update with $\mc{R}(x) = \sum_{i=1}^n x_i \log(x_i)$ as the function that is strongly convex w.r.t. $\norm{\cdot}_1$. Correspondingly, $\nabla_t = f_t^\top A_t$ and $M_t = f_{t-1}^{\top} A_{t-1}$. Following the line of proof in Lemma \ref{Sasha-Karthik}, in particular, using Equation \ref{eq:aistats} for the specific case with $\mathcal{D}_{\mc{R}}$ as KL divergence, we get that for any $t$ and any $u_t \in \Delta_n$,
\begin{align*}
f_t^\top A_t x_t -  f_t^\top A_t u_t &\leq \frac{1}{\eta_t}\bigg\{\sum_{i=1}^{n} u_t[i] \log\left(\frac{\x_{t}[i]}{\x'_{t-1}[i]} \right) - \frac{1}{2}\norm{\hat{x}_{t} - x_t}_1^2 - \frac{1}{2}\norm{\hat{x}'_{t-1} - x_t}_1^2 \bigg\}\\
		&~~~~~~~~~~~~~~~~~~~~~~~~~~+\norm{f_t^\top A_t - f_{t-1}^\top A_{t-1}}_\infty\norm{x_t - \hat{x}_{t}}_1 \\
		&\leq \frac{1}{\eta_t}\bigg\{\sum_{i=1}^{n} u_t[i] \log\left(\frac{\x'_{t}[i]}{\x'_{t-1}[i]} \right) - \frac{1}{2}\norm{\hat{x}_{t} - x_t}_1^2 - \frac{1}{2}\norm{\hat{x}'_{t-1} - x_t}_1^2 \bigg\}\\
		&~~~~~~~~~~~~~~~~~~~~~~~~~~+\norm{f_t^\top A_t - f_{t-1}^\top A_{t-1}}_\infty \norm{x_t - \hat{x}_{t}}_1 + \frac{1}{\eta_t} \max_{i \in [n]} \log\left(\frac{\x_{t}[i]}{\x'_{t}[i]} \right).
\end{align*}
Now let us bound for some $i$ the term, $\log\left(\frac{\x_{t}[i]}{\x'_{t}[i]} \right)$. Notice that if $\x_{t}[i] \le \x'_{t}[i]$ then the term is anyway bounded by 0. Now assume $\x_{t}[i] > \x'_{t}[i]$. Letting $\beta=1/T^2$, since $\x'_t[i] = (1 - T^{-2})\x_t[i] + 1/(n T^2)$, we can have $\x_{t}[i] > \x'_{t}[i]$ only when $\x_{t}[i] > 1/n$. Hence, 
$$
\log\left(\frac{\x_{t}[i]}{\x'_{t}[i]} \right) = \log\left(\frac{\x_{t}[i]}{(1 - T^{-2})\x_t[i] + 1/(n T^2)} \right) \le \frac{2}{T^2}.
$$
Using this we can conclude that :
\begin{align*}
f_t^\top A_t x_t -  f_t^\top A_t u_t &\leq \frac{1}{\eta_t}\bigg\{\sum_{i=1}^n u_t[i] \log\left(\frac{\x'_{t}[i]}{\x'_{t-1}[i]} \right) - \frac{1}{2}\norm{\hat{x}_{t} - x_t}_1^2 - \frac{1}{2}\norm{\hat{x}'_{t-1} - x_t}_1^2 \bigg\}\\
		&~~~~~~~~~~~~~~~~~~~~~~~~~~+\norm{f_t^\top A_t - f_{t-1}^\top A_{t-1}}_\infty \norm{x_t - \hat{x}_{t}}_1 + \frac{2}{T^2} \frac{1}{\eta_t} .
\end{align*}
Summing over $t\in [T]$ we obtain that :
\begin{align*}
\sum_{t=1}^T \left(f_t^\top A_t x_t -  f_t^\top A_t u_t\right) &\leq \sum_{t=1}^T \frac{1}{\eta_t}\bigg\{\sum_{i=1}^n u_t[i] \log\left(\frac{\x'_{t}[i]}{\x'_{t-1}[i]} \right) - \frac{1}{2}\norm{\hat{x}_{t} - x_t}_1^2 - \frac{1}{2}\norm{\hat{x}'_{t-1} - x_t}_1^2 \bigg\}\\
		&~~~~~~~~~~+  \sum_{t=1}^T \norm{f_t^\top A_t - f_{t-1}^\top A_{t-1}}_\infty \norm{x_t - \hat{x}_{t}}_1  + \frac{2}{T^2} \sum_{t=1}^T  \frac{1}{\eta_t} .
\end{align*}
Note that $\frac{1}{\eta_t} \le \mathcal{O}\left(\sqrt{T}\right)$ and so assuming $T$ is large enough,  $\frac{1}{T^2} \sum_{t=1}^T  \frac{1}{\eta_t}  \le 1$ and so,
\begin{align}\label{eq:inter}
\sum_{t=1}^T \left(f_t^\top A_t x_t -  f_t^\top A_t u_t\right) &\leq \sum_{t=1}^T \frac{1}{\eta_t}\bigg\{\sum_{i=1}^n u_t[i] \log\left(\frac{\x'_{t}[i]}{\x'_{t-1}[i]} \right) - \frac{1}{2}\norm{\hat{x}_{t} - x_t}_1^2 - \frac{1}{2}\norm{\hat{x}'_{t-1} - x_t}_1^2 \bigg\} \notag\\
		&~~~~~~~~~~~~~~~~~~~~~~~~~~+  \sum_{t=1}^T \norm{f_t^\top A_t - f_{t-1}^\top A_{t-1}}_\infty \norm{x_t - \hat{x}_{t}}_1   + 1.
\end{align}
Now note that we can rewrite the first sum in the above bound and get :
\begin{align*}
\sum_{t=1}^T \frac{1}{\eta_t} \sum_{i=1}^n u_t[i] \log\left(\frac{\x'_{t}[i]}{\x'_{t-1}[i]} \right)  & \le \sum_{t=2}^T \frac{\sum_{i=1}^n u_t[i] \log\left(\frac{1}{\x'_{t-1}[i]} \right)}{\eta_{t}} - \frac{\sum_{i=1}^n u_{t-1}[i] \log\left(\frac{1}{\x'_{t-1}[i]} \right)}{\eta_{t-1}} + \frac{\log(T^2n)}{\eta_1}\\
& \le \sum_{t=2}^T \frac{\sum_{i=1}^n \left(u_t[i] - u_{t-1}[i] \right) \log\left(\frac{1}{\x'_{t-1}[i]} \right)}{\eta_{t}} \\
& ~~~~~+ \sum_{t=2}^T \sum_{i=1}^n u_{t-1}[i] \log\left(\frac{1}{\x'_{t-1}[i]} \right) \left(\frac{1}{\eta_t} -  \frac{1}{\eta_{t-1}} \right)+ \frac{\log(T^2n)}{\eta_1}.
\intertext{Since by definition of $\x'_{t-1}$, we are mixing in $1/T^2$ of the uniform distribution we have that for any $i$, $\x'_{t-1}[i] > \frac{1}{T^2 n}$ and, since $\eta_t$'s are non-increasing, we continue bounding above as}
\sum_{t=1}^T \frac{1}{\eta_t} \sum_{i=1}^n u_t[i] \log\left(\frac{\x'_{t}[i]}{\x'_{t-1}[i]} \right) & \le \log(T^2 n)\sum_{t=2}^T  \frac{ \norm{u_{t-1} - u_t}_1}{\eta_{t}} + \log(T^2 n)\sum_{t=2}^T  \left(\frac{1}{\eta_t} -  \frac{1}{\eta_{t-1}} \right)+ \frac{\log(T^2n)}{\eta_1} \\
& \le \log(T^2 n)\left( \sum_{t=2}^T  \frac{ \norm{u_{t-1} - u_t}_1}{\eta_{t}} + \frac{1}{\eta_T} - \frac{1}{\eta_1} \right)+ \frac{\log(T^2n)}{\eta_1}\\
& \le \log(T^2 n)\left( \sum_{t=2}^T  \frac{ \norm{u_{t-1} - u_t}_1}{\eta_{t}} + \frac{1}{\eta_T}\right),
\end{align*}
using the above in Equation \ref{eq:inter} we get 
\begin{align*}
\sum_{t=1}^T f_t^\top A_t x_t &-  f_t^\top A_t u_t \\
&\leq \log(T^2 n) \sum_{t=2}^T  \frac{ \norm{u_{t-1} - u_t}_1}{\eta_{t}} - \frac{1}{2} \sum_{t=1}^T\frac{1}{\eta_t} \norm{\hat{x}_{t} - x_t}_1^2 - \frac{1}{2} \sum_{t=1}^T \frac{1}{\eta_t} \norm{\hat{x}'_{t-1} - x_t}_1^2  + 1  \\
		&~~~~~~~~~~~~~~~~~~~+ \sum_{t=1}^T  \norm{f_t^\top A_t - f_{t-1}^\top A_{t-1}}_\infty \norm{x_t - \hat{x}_{t}}_1 + \frac{\log(T^2 n)}{\eta_T} \\
		&\leq \frac{\log(T^2 n)\left( {C}_T (u_1,\ldots,u_T) + 2\right)}{\eta_{T}}  - \frac{1}{2} \sum_{t=1}^T\frac{1}{\eta_t} \norm{\hat{x}_{t} - x_t}_1^2 - \frac{1}{2} \sum_{t=1}^T \frac{1}{\eta_t} \norm{\hat{x}'_{t-1} - x_t}_1^2   \\
		&~~~~~~~~~~~~~~~~~~~+ \sum_{t=1}^T  \norm{f_t^\top A_t - f_{t-1}^\top A_{t-1}}_\infty \norm{x_t - \hat{x}_{t}}_1 . \label{eq:Bbranch} \numberthis
\end{align*}

Notice that our choice of step size given by,
\begin{align*}
\eta_t &= \min\left\{\log(T^2n)\frac{L}{\sqrt{\sum_{i=1}^{t-1}\norm{f_i^\top A_i - f_{i-1}^\top A_{i-1}}^2_\infty } + \sqrt{\sum_{i=1}^{t-2}\norm{f_i^\top A_i - f_{i-1}^\top A_{i-1}}^2_\infty }}, \frac{1}{32 L}\right\}\\
& = \min\left\{\log(T^2n)\frac{L \left(\sqrt{\sum_{i=1}^{t-1}\norm{f_i^\top A_i - f_{i-1}^\top A_{i-1}}^2_\infty } - \sqrt{\sum_{i=1}^{t-2} \norm{f_i^\top A_i - f_{i-1}^\top A_{i-1}}^2_\infty }\right)}{\norm{f_{t-1}^\top A_{t-1} - f_{t-2}^\top A_{t-2}}^2_\infty }, \frac{1}{32 L}\right\}, \numberthis \label{ssssstep}
\end{align*}
guarantees that 
\begin{align*}
\eta^{-1}_t = \max\left\{\frac{\sqrt{\sum_{i=1}^{t-1}\norm{f_i^\top A_i - f_{i-1}^\top A_{i-1}}^2_\infty } + \sqrt{\sum_{i=1}^{t-2}\norm{f_i^\top A_i - f_{i-1}^\top A_{i-1}}^2_\infty }}{\log(T^2n) L}, 32 L\right\}.
\end{align*}
Using the step-size specified above in the bound \ref{eq:Bbranch}, we get 
\begin{align*}
\sum_{t=1}^T f_t^\top &A_t x_t - \sum_{t=1}^T f_t^\top A_t u_t \\
&\leq  \log(T^2 n)\left( {C}_T (u_1,\ldots,u_T) + 2\right) \left(\frac{2\sqrt{\sum_{t=1}^T \norm{f_t^\top A_t - f_{t-1}^\top A_{t-1}}^2_\infty} }{\log(T^2n) L}+32L \right)  \\
		&+ \sum_{t=1}^T  \norm{f_t^\top A_t - f_{t-1}^\top A_{t-1}}_\infty \norm{x_t - \hat{x}_{t}}_1- 16 L  \sum_{t=1}^T\norm{\hat{x}_{t} - x_t}_1^2 - 16 L \sum_{t=1}^T  \norm{\hat{x}'_{t-1} - x_t}_1^2. \label{eq:234} \numberthis
\end{align*}
Now note that by triangle inequality, we have 
\begin{align*}
 \norm{f_t^\top A_t - f_{t-1}^\top A_{t-1}}_\infty  &=  \norm{f_t^\top A_t -f_t^{\top}A_{t-1} + f^\top_tA_{t-1} - f_{t-1}^\top A_{t-1} }_\infty \\
& \le  \norm{A_{t-1} - A_t}_\infty +  \norm{f_t - f_{t-1}}_1\\
& \le  \norm{A_{t-1} - A_t}_\infty + \norm{f_t  - \hat{f}_{t-1}}_1 + \norm{\hat{f}_{t-1} - f_{t-1}}_1, 
\end{align*}
since the entries of matrix sequence $\{A_t\}_{t=1}^T$ are bounded by one. Using the bound above in \eqref{eq:234} and splitting the product term, we see that {\small
\begin{align*}
\sum_{t=1}^T \left(f_t^\top A_t x_t -  f_t^\top A_t u_t \right)&\leq  \log(T^2 n)\left( {C}_T (u_1,\ldots,u_T) + 2\right) \left(\frac{2\sqrt{\sum_{t=1}^T \norm{f_t^\top A_t - f_{t-1}^\top A_{t-1}}^2_\infty} }{\log(T^2n) L}+32L \right)  \\
		&+ 2\sum_{t=1}^T  \norm{ A_t -  A_{t-1}}_\infty - 8 L  \sum_{t=1}^T\norm{\hat{x}_{t} - x_t}_1^2 - 16 L \sum_{t=1}^T  \norm{\hat{x}'_{t-1} - x_t}_1^2\\
		&+\frac{1}{16L}  \sum_{t=1}^T \norm{f_t  - \hat{f}_{t-1}}^2_1 +\frac{1}{16L}  \sum_{t=1}^T  \norm{\hat{f}_{t-1} - f_{t-1}}_1^2, \numberthis \label{eqeqeq}
\end{align*}}
where we used the simple inequality $ab \leq \frac{\rho}{2}a^2+\frac{1}{2\rho}b^2$ for $\rho> 0$. 
\paragraph{When Player II follows prescribed strategy}
In this case we would like to get convergence of payoffs to the average value of the games. To get this, using the notation $x^*_t = \argmin{x_t \in \Delta_n}  f_t^\top A_t x_t$ and denoting the corresponding sequence regularity for Player I by $C_T$, we get {\small
\begin{align*}
\sum_{t=1}^T \left(f_t^\top A_t x_t -  f_t^\top A_t x^*_t \right)&\leq  \log(T^2 n)\left( {C}_T + 2\right) \left(\frac{2\sqrt{\sum_{t=1}^T \norm{f_t^\top A_t - f_{t-1}^\top A_{t-1}}^2_\infty} }{\log(T^2n)L}+32L \right)  \\
		& +2\sum_{t=1}^T  \norm{ A_t -  A_{t-1}}_\infty - 8 L  \sum_{t=1}^T\norm{\hat{x}_{t} - x_t}_1^2 - 16 L \sum_{t=1}^T  \norm{\hat{x}'_{t-1} - x_t}_1^2\\
		&+\frac{1}{16L}  \sum_{t=1}^T \norm{f_t  - \hat{f}_{t-1}}^2_1 +\frac{1}{16L}  \sum_{t=1}^T  \norm{\hat{f}_{t} - f_{t}}_1^2+\frac{1}{4L},
\end{align*}}
where the term $\frac{1}{4L}$ appeared in the last line comparing to \eqref{eqeqeq} is due to
\begin{align*}
\frac{1}{16L}  \sum_{t=1}^T  \norm{\hat{f}_{t-1} - f_{t-1}}_1^2-\frac{1}{16L}  \sum_{t=1}^T  \norm{\hat{f}_{t} - f_{t}}_1^2 \leq \frac{1}{4L}.
\end{align*}
Using the same bound for Player 2 (using loss as $- f_t^\top A_t x_t$ on round $t$), as well as using $f^*_t = \argmin{f_t \in \Delta_m} - f_t^\top A_t x_t$ and denoting the corresponding sequence regularity by $C'_T$, we have that {\small
\begin{align*}
\sum_{t=1}^T \left(f_t^\top A_t x_t -  {f^*_t}^\top A_t x_t \right)&\geq   -\log(T^2 m)\left( {C}'_T + 2\right) \left(\frac{2\sqrt{\sum_{t=1}^T \norm{ A_tx_t - A_{t-1}x_{t-1}}^2_\infty }}{\log(T^2m)L}+32L \right)  \\
		&-2 \sum_{t=1}^T  \norm{ A_t -  A_{t-1}}_\infty + 8 L  \sum_{t=1}^T\norm{\hat{f}_{t} - f_t}_1^2 + 16 L \sum_{t=1}^T  \norm{\hat{f}'_{t-1} - f_t}_1^2\\
		&-\frac{1}{16L}  \sum_{t=1}^T \norm{x_t  - \hat{x}_{t-1}}^2_1 -\frac{1}{16L}  \sum_{t=1}^T  \norm{\hat{x}_{t} - x_{t}}_1^2-\frac{1}{4L}.
\end{align*}}
Combining the two and noting that 
\begin{align*}
{f^*_t}^\top A_t x_t = \sup_{f_t \in \Delta_m}f_t^\top A_t x_t &\ge \inf_{x_t \in \Delta_n} \sup_{f_t \in \Delta_m} f_t^\top A_t x_t \\
&= \sup_{f_t \in \Delta_m} \inf_{x_t \in \Delta_n} f_t^\top A_t x_t \ge \inf_{x_t \in \Delta_n} f^\top_t A_t x_t = f^\top_t A_t x^*_t ,
\end{align*}
we get 
\begin{align*}
\sum_{t=1}^T \sup_{f_t \in \Delta_m} f_t^\top A_t x_t  &\le \sum_{t=1}^T \inf_{x_t \in \Delta_n}\sup_{f_t \in \Delta_m} f_t^\top A_t x_t +\frac{256L}{T}+\frac{1}{2L}+4 \sum_{t=1}^T  \norm{ A_t -  A_{t-1}}_\infty\\
&+\log(T^2 n)\left( {C}_T + 2\right) \left(\frac{2\sqrt{\sum_{t=1}^T \norm{f_t^\top A_t - f_{t-1}^\top A_{t-1}}^2_\infty} }{\log(T^2n)L}+32L \right) \\
&+\log(T^2 m)\left( {C}'_T + 2\right) \left(\frac{2\sqrt{\sum_{t=1}^T \norm{ A_tx_t - A_{t-1}x_{t-1}}^2_\infty }}{\log(T^2m)L}+32L \right)\\
& +\left(\frac{1}{16L}- 8L\right)   \sum_{t=1}^T \norm{\hat{x}_{t} - x_t}_1^2 +\left(\frac{1}{16L}- 16L\right)  \sum_{t=1}^T  \norm{\hat{x}_{t-1} - x_t}_1^2  \\
&+\left(\frac{1}{16L}- 8L\right) \sum_{t=1}^T\norm{\hat{f}_{t} - f_t}_1^2 +\left(\frac{1}{16L}- 16L\right) \sum_{t=1}^T  \norm{\hat{f}_{t-1} - f_t}_1^2  ,\label{eq:gm} \numberthis
\end{align*}
where the constant $256L/T$ appeared in the first line accounts for the identities
\begin{align*}
\norm{\hat{x}_{t-1} - x_t}_1^2-\norm{\hat{x}'_{t-1} - x_t}_1^2 \leq \frac{8}{T^2}\ \ \ \ \ \ \ \norm{\hat{f}_{t-1} - f_t}_1^2-\norm{\hat{f}'_{t-1} - f_t}_1^2 \leq \frac{8}{T^2}.
\end{align*}
Using the triangle inequality again, 
\begin{align*}
\sum_{t=1}^T \norm{f_t^\top A_t - f_{t-1}^\top A_{t-1}}^2_\infty  &= \sum_{t=1}^T \norm{f_t^\top A_t -f_t^{\top}A_{t-1} + f_t^\top A_{t-1} - f_{t-1}^\top A_{t-1}}^2_\infty \\
& \le 2 \sum_{t=1}^T\norm{A_{t-1} - A_t}_\infty^2 + 2 \sum_{t=1}^T\norm{f_t - f_{t-1}}_1^2\\
& \le 2 \sum_{t=1}^T\norm{A_{t-1} - A_t}_\infty^2 + 4\sum_{t=1}^T\norm{f_t  - \hat{f}_{t-1}}_1^2 + 4\sum_{t=1}^T\norm{\hat{f}_{t-1} - f_{t-1}}_1^2, \numberthis \label{eq:12345}
\end{align*}
which also implies {\small
\begin{align*}
\sqrt{\sum_{t=1}^T \norm{f_t^\top A_t - f_{t-1}^\top A_{t-1}}^2_\infty} &\leq  \sqrt{2 \sum_{t=1}^T\norm{A_{t-1} - A_t}_\infty^2 + 4\sum_{t=1}^T\norm{f_t  - \hat{f}_{t-1}}_1^2 + 4\sum_{t=1}^T\norm{\hat{f}_{t-1} - f_{t-1}}_1^2}\\
&\leq2\sqrt{ \sum_{t=1}^T\norm{A_{t-1} - A_t}_\infty^2 }+ 2\sqrt{ \sum_{t=1}^T\norm{f_t  - \hat{f}_{t-1}}_1^2 + \sum_{t=1}^T\norm{\hat{f}_{t-1} - f_{t-1}}_1^2}\\
&\leq2\sqrt{ \sum_{t=1}^T\norm{A_{t-1} - A_t}_\infty^2 }+2+2\sum_{t=1}^T\norm{f_t  - \hat{f}_{t-1}}_1^2 + 2\sum_{t=1}^T\norm{\hat{f}_{t-1} - f_{t-1}}_1^2\\
&\leq2\sqrt{ \sum_{t=1}^T\norm{A_{t-1} - A_t}_\infty^2 }+10+2\sum_{t=1}^T\norm{f_t  - \hat{f}_{t-1}}_1^2 + 2\sum_{t=1}^T\norm{\hat{f}_{t} - f_{t}}_1^2 \numberthis \label{eq:123456},
\end{align*}}
where we used the bound $\sqrt{c} \leq c+1$ for any $c \geq 0$ in the penultimate line. Similar bounds as Equations \eqref{eq:12345} and \eqref{eq:123456} hold for the other player as well. Using them in Equation \ref{eq:gm} after some calculations, we conclude that 
{\small \begin{align*}
\sum_{t=1}^T \sup_{f_t \in \Delta_m} f_t^\top A_t x_t & \le \sum_{t=1}^T \inf_{x_t \in \Delta_n}\sup_{f_t \in \Delta_m} f_t^\top A_t x_t +\frac{256L}{T}+\frac{1}{2L}+4\sum_{t=1}^T\norm{A_{t-1} - A_t}_\infty\\
& + 32L\big( \log(T^2 n) {C}_T + \log(T^2 m) C'_T  + 2 \log(T^4 n m)   \big) +\left(C_T+C'_T+4\right)\frac{20 + 4 \sqrt{\sum_{t=1}^T\norm{A_{t-1} - A_t}_\infty^2}}{L}\\ 
& +   4\left(\frac{   {C}_T  +3}{L}  - 2L  \right) \left( \sum_{t=1}^T\norm{\hat{f}_{t} - f_t}_1^2 + 2 \sum_{t=1}^T  \norm{\hat{f}_{t-1} - f_t}_1^2 \right)\\
& +  4 \left(\frac{   {C}'_T  + 3}{L}  - 2L  \right) \left( \sum_{t=1}^T \norm{\hat{x}_{t} - x_t}_1^2 + 2  \sum_{t=1}^T  \norm{\hat{x}_{t-1} - x_t}_1^2\right).
\end{align*}}

\paragraph{When Player II is dishonest}
In this case we would like to bound Player I's regret regardless of the strategy adopted by Player II. Dropping one of the negative terms in Equation \ref{eq:Bbranch}, we get : 
\begin{align*}
\sum_{t=1}^T \left(f_t^\top A_t x_t -  f_t^\top A_t u_t\right) 		&\leq \frac{\log(T^2 n)\left( {C}_T (u_1,\ldots,u_T) + 2\right)}{\eta_{T}}  - \frac{1}{2} \sum_{t=1}^T\frac{1}{\eta_t} \norm{\hat{x}_{t} - x_t}_1^2 \\
		&+ \sum_{t=1}^T  \norm{f_t^\top A_t - f_{t-1}^\top A_{t-1}}_\infty \norm{x_t - \hat{x}_{t}}_1 \\
		&\leq \frac{\log(T^2 n)\left( {C}_T (u_1,\ldots,u_T) + 2\right)}{\eta_{T}}  - \frac{1}{2} \sum_{t=1}^T\frac{1}{\eta_t} \norm{\hat{x}_{t} - x_t}_1^2 \\
		&+ \sum_{t=1}^T  \frac{\eta_{t+1}}{2}\norm{f_t^\top A_t - f_{t-1}^\top A_{t-1}}^2_\infty + \frac{1}{2}\sum_{t=1}^T \frac{1}{\eta_{t+1}}\norm{x_t - \hat{x}_{t}}^2_1 .  \numberthis \label{1:48am}
\end{align*}
Noting to the telescoping sum
\begin{align*}
\frac{1}{2}\sum_{t=1}^T \left(\frac{1}{\eta_{t+1}}-\frac{1}{\eta_{t}}\right)\norm{x_t - \hat{x}_{t}}^2_1 \leq 2\sum_{t=1}^T \left(\frac{1}{\eta_{t+1}}-\frac{1}{\eta_{t}}\right) \leq \frac{2}{\eta_{T+1}},
\end{align*}
as well as the choice of step-size \eqref{ssssstep} which entails
\begin{align*}
\sum_{t=1}^T  \frac{\eta_{t+1}}{2}\norm{f_t^\top A_t - f_{t-1}^\top A_{t-1}}^2_\infty &\leq \log(T^2n)\frac{L}{2} \sum_{t=1}^T \sqrt{\sum_{i=1}^{t}\norm{f_i^\top A_i - f_{i-1}^\top A_{i-1}}^2_\infty } - \sqrt{\sum_{i=1}^{t-1} \norm{f_i^\top A_i - f_{i-1}^\top A_{i-1}}^2_\infty }\\
&\leq \log(T^2n)\frac{L}{2}\sqrt{\sum_{t=1}^{T}\norm{f_t^\top A_t - f_{t-1}^\top A_{t-1}}^2_\infty },
\end{align*}
we bound \eqref{1:48am} to obtain
\begin{align*}
\sum_{t=1}^T \left(f_t^\top A_t x_t -  f_t^\top A_t u_t\right) &\leq \frac{\log(T^2 n)\left( {C}_T (u_1,\ldots,u_T) + 2\right)}{\eta_{T}}+ \frac{2}{\eta_{T+1}}+ \log(T^2n)\frac{L}{2}\sqrt{\sum_{t=1}^{T}\norm{f_t^\top A_t - f_{t-1}^\top A_{t-1}}^2_\infty }\\
&\leq 2\log(T^2 n)\left( {C}_T (u_1,\ldots,u_T) + 2\right)\left(32L+\frac{2\sqrt{\sum_{t=1}^{T}\norm{f_t^\top A_t - f_{t-1}^\top A_{t-1}}^2_\infty }}{\log(T^2n)L}\right)\\
&+ \log(T^2n)\frac{L}{2}\sqrt{\sum_{t=1}^{T}\norm{f_t^\top A_t - f_{t-1}^\top A_{t-1}}^2_\infty }.
\end{align*}
A similar statement holds for Player II that her/his pay off converges at the provided rate to the average minimax equilibrium value.$\hfill \blacksquare $\\

\bibliographystyle{IEEEtran}
\bibliography{IEEEabrv,shahin}